\pdfoutput=1

\documentclass[11pt]{article}

\usepackage{listings}
\usepackage[most]{tcolorbox}
\tcbuselibrary{listings, skins} 
\usepackage[utf8]{inputenc}
\usepackage{textgreek}
\usepackage{xcolor}
\usepackage{amssymb}
\usepackage{float}

\definecolor{myblue}{rgb}{0.2, 0.2, 0.8}
\definecolor{mygreen}{rgb}{0.1, 0.6, 0.1}
\definecolor{myorange}{rgb}{0.9, 0.4, 0.0}

\newcommand{\casesize}{\fontsize{8.5pt}{10.5pt}\selectfont}

\newtcolorbox{promptbox}[1][]{
  colback=white!96!brown,
  colframe=brown!40!black,
  boxrule=1pt,
  arc=4pt,
  boxsep=5pt,
  left=12pt,
  enhanced,
  breakable,
  drop shadow={brown!30!black, opacity=0.35},
  title=#1,
  fonttitle=\bfseries\color{brown!50!black},
  colbacktitle=brown!8!white,
  attach boxed title to top left={yshift=-0.25em, xshift=1.4em},
  boxed title style={
    arc=2.5pt,
    boxrule=0.75pt,
  },
  listing only,
  listing options={
    language=Python,
    basicstyle=\small\ttfamily,
    breaklines=true,
    breakatwhitespace=false,
    postbreak=\mbox{\textcolor{brown!60!black}{\tiny$\hookrightarrow$}\space},
    showstringspaces=false,
    frame=none,
    tabsize=2,
    commentstyle=\color{brown!60!black},
    keywordstyle=\color{brown!80!black},
    stringstyle=\color{red!60!black},
    literate=
      *{f"}{f"}{1}
      *{'}{'}{1}
      {\{}{{\{}}1
      {\}}{{\}}}1,
  },
}

\usepackage[final]{acl}

\usepackage{times}
\usepackage{latexsym}

\usepackage[T1]{fontenc}

\usepackage[utf8]{inputenc}

\usepackage{microtype}

\usepackage{inconsolata}

\usepackage{enumitem}
\usepackage{graphicx}
\usepackage{listings}
\usepackage{booktabs}
\usepackage{multirow}
\usepackage[table]{xcolor}
\usepackage{colortbl}
\usepackage{amsmath}
\usepackage{xcolor}
\usepackage{makecell}
\usepackage{tabularx}
\usepackage{booktabs} 

\usepackage{array} 
\renewcommand{\arraystretch}{1} 

\title{ThinkPilot: Steering Reasoning Models  via  \\ Automated Think-prefixes Optimization }

\author{
  \textbf{Sunzhu Li}$^{1}$\textbf{,}
  \textbf{Zhiyu Lin}$^2$ \textbf{,}
  \textbf{Shuling Yang}$^1$\textbf{,} 
  \textbf{Jiale Zhao}$^1$\textbf{,} \\
  \textbf{Wei Chen}$^1$ \thanks{\ \ Corresponding Author}  \\
  $^{1}$Li Auto Inc., China\\
  $^{2}$The Chinese University of Hong Kong, Shenzhen, China \\
  {\tt 
        \{lisunzhu, yangshuling, zhaojiale5, chenwei10\}@lixiang.com
  }\\
  {\tt 
    zhiyulin1@link.cuhk.edu.cn
  }
}

\begin{document}
\maketitle
\begin{abstract}

Large Reasoning Models (LRMs) are powerful, but they still suffer from inefficient and off-target reasoning. Currently, training-free methods are limited to either rigid heuristics or descriptive, non-actionable analyses. In this paper, we introduce ThinkPilot, a training-free framework that automatically optimizes LRMs reasoning. It uses an evolutionary process to generate \textit{think-prefixes}, namely instructions that evolve driven by a taxonomy of \textit{reasoning behaviors} to guide models toward superior performance. Extensive experiments demonstrate ThinkPilot's broad effectiveness: it significantly improves the accuracy-length trade-off for efficient reasoning, drastically improves safety (e.g., cutting the StrongREJECT score of DeepSeek-R1-Distill-Qwen-32B from 27.0\% to 0.7\%), and enhances instruction following. It also synergizes with existing training-based methods. Specially, our analysis reveals that think-prefixes can reliably control LRMs’ reasoning behaviors,  and that different tasks have strong preferences for specific behavioral distributions. By automatically identifying and eliciting these behaviors, ThinkPilot provides a generalizable framework for aligning LRMs reasoning with task demands.

\end{abstract}

\begin{figure*}[htbp]
  \centering
  \includegraphics[width=0.9\textwidth]{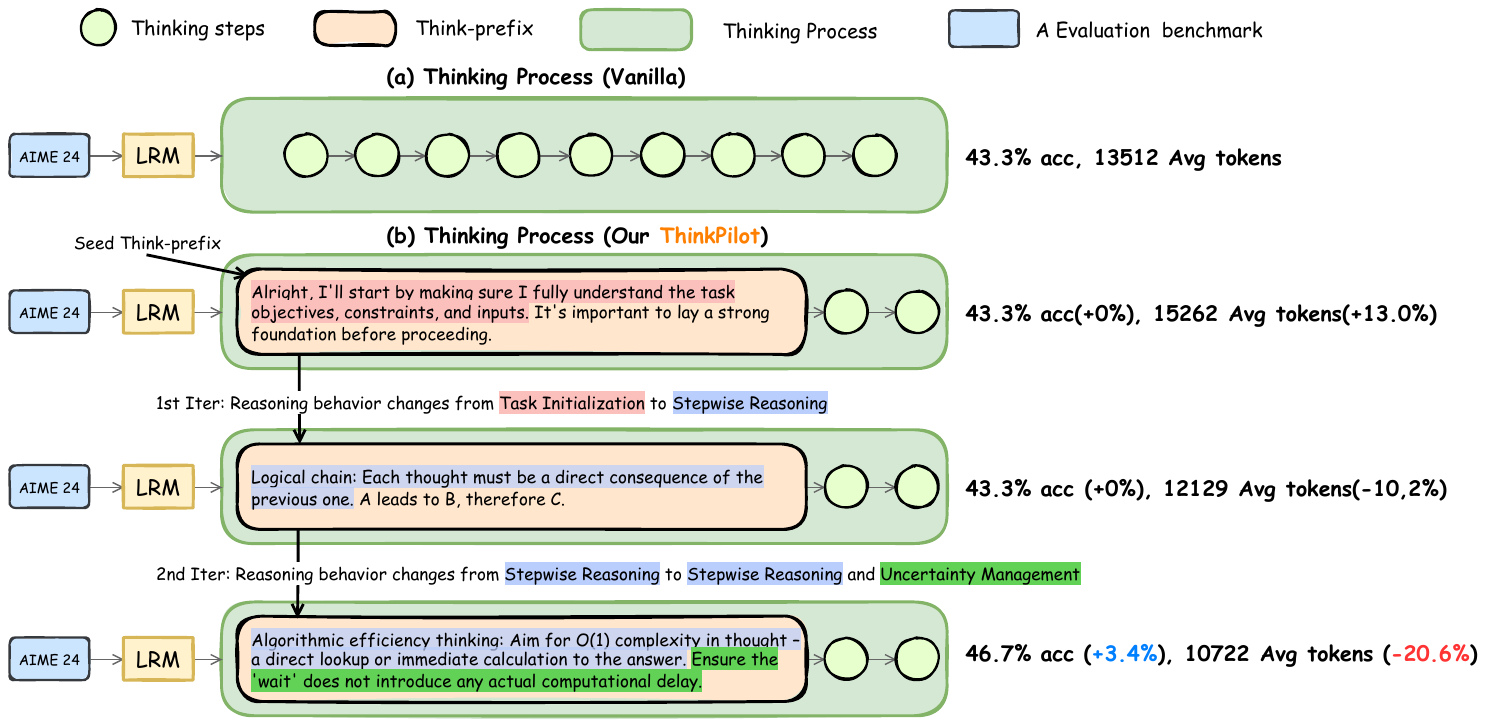}
  \caption{\textbf{The comparison between (a) vanilla thinking process and (b) ThinkPilot}, which  guides an LRM by iteratively optimizing think-prefixes based on reasoning behaviors. On the R1-Qwen-7B model, after two iterations, ThinkPilot improves accuracy by 3.4\% while reducing average token usage by 20.6\% on AIME 24. }
  \label{intro}
\end{figure*}

\section{Introduction}

Large Reasoning Models (LRMs) \citep{jaech2024openai, guo2025deepseek} have achieved notable progress in complex tasks like math problem solving and code generation. These models support iterative thinking and better problem decomposition by generating detailed reasoning before final answers \citep{chen2025towards}. However, LRMs still face issues such as overly lengthy reasoning, and off-target responses that deviate from instructions or expectations, which wastes computation and harms answer quality  \citep{chen2024not, cuadron2025danger, gan2025rethinking}. Thus, to  improve performance, guiding LRMs toward more efficient and task-aligned  reasoning patterns is essential.

To address these issues, existing efforts fall into two main categories. \textit{Training-based} approaches adjust model parameters via supervised fine-tuning or reinforcement learning to encourage behaviors like safety or efficiency \citep{ma2025cot, aggarwal2025l1, chen2025towards}, but they require expensive supervision or task-specific reward design. In contrast, \textit{training-free} methods steer reasoning without changing model weights, offering greater flexibility and scalability, which makes them especially attractive for practical deployment. Given these advantages, we focus on recent advances and challenges in \textit{training-free} ones.

Among training-free methods, current research can be divided into two primary categories, each with notable limitations. First, \textbf{human-heuristic methods} \citep{wu2025effectively, ma2025reasoning, wang2025wait, handelman2009thought} guide the reasoning process by injecting human-crafted phrases to make it more compact or safer. However, these heuristics often lack principled theoretical grounding, making them difficult to generalize across tasks and models. Second, \textbf{interpretability-driven analysis } \citep{wang2025beyond, ghosal2025does, zhang2025no, ma2025reasoning, wu2025effectively} has turned to understand the reasoning processes, such as assessing the importance of words or sentences within the reasoning paths. Yet, the efforts tend to remain descriptive, rarely yielding actionable strategies for model intervention. Naturally, these limitations raise a fundamental question: \textit{can we develop an automatic and interpretability-driven framework, to efficiently discover the reasoning interventions for LRMs?}

In this paper, we introduce \textbf{ThinkPilot}, a training-free method that optimizes LRMs performance by strategically and automatically guiding their reasoning process. Inspired by Schoenfeld’s Episode Theory of human problem solving~\citep{schoenfeld2014mathematical} and recent studies~\citep{bogdan2025thought,wang2025beyond,venhoff2025understanding,li2025understanding} on structured model reasoning, ThinkPilot introduces a principled taxonomy that defines \textit{reasoning behaviors}—i.e., the observable and controllable strategies adopted by models during  thinking. 
Built upon this taxonomy, ThinkPilot uses an evolution-inspired workflow to discover effective reasoning interventions. Specifically, it  generates \textit{think-prefixes}, which are interventional instructions inserted at the start of the thinking process, to trigger desired reasoning. Through iterative refinement, these prefixes gradually shift the reasoning behaviors they control, thereby enabling the identification of task-preferred reasoning behaviors and achieving superior performance. As shown in Figure ~\ref{intro} , ThinkPilot achieves desired performance with significantly lower token overhead compared to vanilla approaches.

Experimental results show that ThinkPilot demonstrates \textit{broad effectiveness across diverse tasks}. In Efficient Reasoning, it significantly improves the model's accuracy-length trade-off, achieving higher accuracy with more concise outputs than baseline methods.  its impact on Safety is particularly notable: it reduced the StrongREJECT score of R1-Qwen-32B from 27.0\% to just 0.7\%, without compromising other reasoning abilities. In Instruction Following, it boosted the IFEval score of R1-Qwen-32B by 6.4 points. Furthermore, consistent gains are  observed on the logical reasoning benchmark BBH. Crucially, ThinkPilot also \textit{synergizes with training-based methods}, further reducing SAFECHAIN's StrongREJECT score from an already low 19.4\% to just 1.4\%.

\begin{figure*}[htbp]
  \centering
  \includegraphics[width=0.9\textwidth]{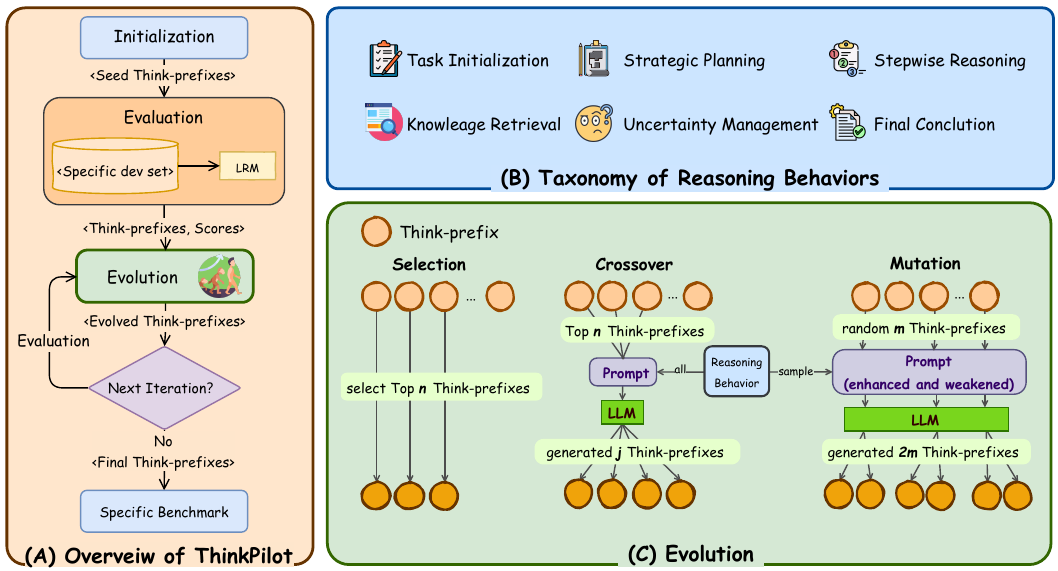}
  \caption{\textbf{Overview of the ThinkPilot}. The method optimizes think-prefixes through an evolutionary loop (A), where the evolution process (C) involves selection, crossover, and mutation, guided by the taxonomy of  reasoning behaviors (B). The complete prompts for crossover and mutation can refer to Appendix~\ref{app:detailed prompt}.  }
  \label{main2}
\end{figure*}

To further understand the source of ThinkPilot’s performance gains, we conducted analysis and identified two key insights. First, existing studies show that LRMs may fail to follow the  instructions for controlling thinking process \citep{wu2025effectively}, whereas we demonstrate that \textit{think-prefixes can reliably and precisely control reasoning behaviors for LRMs}. This enables LRMs to be steered in desired directions. Second, \textit{different tasks favor distinct reasoning behaviors, and this preference strongly correlated to performance}. For example, behaviors that are helpful in some tasks may be ineffective or even harmful in others. This reveals the importance of aligning behavior strategies with task characteristics. In practice, ThinkPilot automatically identifies and elicits the behaviors each task prefers, thus ultimately leading to the performance that align with human expectations. In summary, our contributions are as follows:

\begin{itemize}

\item We propose ThinkPilot, a novel  training-free framework that uses an evolutionary algorithm guided by a taxonomy of reasoning behaviors, automatically discovering the think-prefixes for steering model reasoning.

\item We reveal that think-prefixes enable precise control of LRMs’ reasoning behaviors, and that aligning these behaviors with task-specific preferences boosts performance.

\item Extensive experiments validate that ThinkPilot as a highly effective and general framework, significantly improves efficiency, safety, and instruction-following capabilities. Moreover, it can effectively synergize with existing training-based methods.

\end{itemize}

\section{ThinkPilot}
In this section, we introduce the workflow of \textit{ThinkPilot}. As shown in Figure~\ref{main2}, the framework operates in two stages: an initialization–evaluation phase  constructs and assesses seed think-prefixes, and an \textbf{evolution–iteration} phase. The latter combines \textit{reasoning behavior modeling} (Figure~\ref{main2} (B)) with evolutionary strategies (\textit{selection}, \textit{mutation}, and \textit{crossover}) to iteratively refine think-prefixes (Figure~\ref{main2} (C)) based on performance feedback, producing high-quality ones for downstream tasks.

\subsection{Initialization and Evaluation}

To initiate the iterative process, we automatically generate seed think-prefixes for each task using an LLM guided by prompts. This prompt-based generation produces diverse candidates varying in control strength, narrative style, and length, enriching the initial search space.  The LLM generates seeds in a first-person perspective to emulate the model’s internal monologue and align them with task objectives (e.g., using “\texttt{Ok, let’s think concisely.}” to encourage brevity). In addition, we find that the optimization process is largely insensitive to the number or quality of the seeds—thus, even a small set of LLM-generated prefixes can effectively bootstrap the process (see Appendix~\ref{app:initial seeds}). Each resulting candidate is then evaluated on the downstream task using metrics such as accuracy or safety, providing feedback to guide further optimization.

\subsection{Evolution and Iteration}

The \textit{Evolution and Iteration} phase optimizes think-prefixes using an evolutionary algorithm (EA) suited to the discrete, non-differentiable search space of natural language. Unlike gradient-based methods, EAs efficiently explore prompts as black-box optimizers~\citep{guo2025evoprompt,fernando2023promptbreeder}. Guided by validation scores from the previous stage, the algorithm iteratively applies mutation, crossover, and selection to evolve new candidates (Figure~\ref{main2} (C)). Our EA embeds the taxonomy of reasoning behaviors into these operations, enabling semantically guided rather than blind exploration, and repeats until convergence to yield prefixes optimized for the target LRM and task.

\newcolumntype{L}{>{\raggedright\arraybackslash}X}
\newcolumntype{S}{>{\raggedright\arraybackslash}p{2cm}} 
\newcolumntype{T}{>{\raggedright\arraybackslash}p{3.5cm}} 

\begin{table*}[t]
\small
\centering
\renewcommand{\arraystretch}{1.}
\begin{tabularx}{\linewidth}{T L L S}
\toprule
\textbf{Types} & \textbf{Definition} & \textbf{Example} & \textbf{Episode} \\
\midrule
\textbf{Task Initialization} &
In the initial reasoning phase, the model identifies its task objectives, constraints, and inputs. &
“Okay, I need to …”,\newline “My task is to …” &
Reading, Analyzing  \\
\textbf{Strategic Planning} &
Before execution, explicitly state or determine a structured action plan or strategic blueprint. &
“I will first …, then …”,\newline “To solve this, I’ll …” &
Planning \\
\textbf{Knowledge Retrieval} &
Review relevant knowledge for problem-solving. &
“According to my knowledge …” &
Analyzing, Planning \\
\textbf{Stepwise Reasoning} &
Execute independent reasoning or computation steps based on the planned logic. &
“… So”, “… Therefore”, \newline“… First” &
Executing \\
\textbf{Uncertainty Management} &
The model pauses and flags confusion or uncertainty when encountering ambiguity. &
“Wait …”, “Hmm …”, “Well …”, “Actually …” &
Monitoring, Evaluating \\
\textbf{Final Conclusion} &
Present the final conclusion. &
“In conclusion …” &
Evaluating, Summarizing \\
\bottomrule
\end{tabularx}
\caption{\textbf{Taxonomy of reasoning behaviors} with definitions and examples, aligned with \textit{Schoenfeld’s Episode Theory} of human problem solving~\citep{schoenfeld2014mathematical}.}
\label{tab:taxonomy_reasoning}
\end{table*}

\paragraph{Taxonomy of Reasoning Behavior} 
Our taxonomy is inspired by \textit{Schoenfeld’s Episode Theory }of human problem solving~\citep{schoenfeld2014mathematical}, which posits that reasoning unfolds through cognitively distinct episodes such as \textit{Reading, Analyzing, Planning, Executing, Monitoring, and Evaluating}. It is also informed by recent studies~\citep{bogdan2025thought,wang2025beyond,venhoff2025understanding,li2025understanding} that model reasoning as a structured, stage-wise cognitive process in LRMs.

Building on these theoretical and empirical foundations, we map the typical reasoning episodes to six atomic behaviors frequently observed in LRMs (Table~\ref{tab:taxonomy_reasoning}). This correspondence suggests that our taxonomy is not heuristically crafted but reflects stable reasoning patterns that emerge across diverse tasks and models, forming a \textit{compact yet extensible behavioral basis} for controllable reasoning. We integrate this taxonomy as prior knowledge into our evolutionary process to guide the generation of effective think-prefixes.

\paragraph{Selection}

\textit{Selection} preserves the top-scoring think-prefixes from the previous evaluation. The best $\textit{n}$ prefixes advance to the next round, preventing strong candidates from being lost and providing a solid basis for subsequent mutation and crossover, thereby ensuring steady evolutionary progress.

\paragraph{Crossover} 
\textit{Crossover}  aims to generate novel think-prefixes by synthesizing complementary reasoning behaviors from the top $n$ performing prefixes. The process leverages a Large Language Model (LLM), such as GPT-4o, guided by a few-shot prompt. Specifically, we select the top-$n$ think-prefixes, denoted as $\{s_i\}_{i=1}^{n}$. These prefixes, along with a reasoning behavior classification ($RB$), are formatted into a tailored prompt, $\textit{Prompt}_{\text{crossover}}(\cdot)$. This prompt instructs the LLM to analyze the behaviors within $\{s_i\}$ and generate  $j$ new prefixes that effectively blend these complementary behaviors. The entire generation process can be formalized as follows:
\begin{equation}\small
\label{Crossover}
\begin{aligned}
\{c_i\}_{i=1}^{j} = \text{LLM}\Big(\textit{Prompt}_{\text{crossover}}\big(\{s_i\}_{i=1}^n , RB ,j\big)\Big)
\end{aligned}
\end{equation}

where ${c_i}_{i=1}^{j}$ are the resulting think-prefixes.

\paragraph{Mutation} 
\textit{Mutation} introduces targeted perturbations, guided by specific reasoning behaviors, to enhance population diversity and explore potentially superior think-prefixes. The process begins by randomly selecting $m$ think-prefixes from the current population. For each selected prefix $s$, we independently assign a randomly chosen reasoning behavior $rb$ (e.g., task initialization, strategic planning) to guide its transformation. Given that the optimal influence of a reasoning behavior may vary across tasks, we introduce two directional perturbations: \textit{enhanced}, which amplifies the influence of the assigned behavior, and \textit{weakened}, which reduces it. This bidirectional mechanism broadens the exploration of compatibility between think-prefixes and reasoning behaviors, thereby enhancing the effectiveness of the mutation operator.

To implement this, we designed a mutation prompt template, 
$\textit{Prompt}_{\text{mutation}}(\cdot)$. For each of the 
$m$ pairs of a prefix $s$ and its assigned reasoning behavior $rb$, we use this template to construct a prompt. This prompt instructs an instruction-following LLM to  simultaneously generate two new think-prefixes: one \textit{enhanced} version and one \textit{weakened} version.  Thus, $m$ calls to the LLM produce a total of $2m$ new candidate think-prefixes. This mutation process for a single prefix $s$ can be formalized as:
\begin{equation}\small
\label{Mutation}
\begin{aligned}
{s_{\text{enhanced}}, s_{\text{weakened}}} = \text{LLM}\Big(\textit{Prompt}_{\text{mutation}}\big(s, rb\big)\Big)
\end{aligned}
\end{equation}

\begin{table*}[!t]
\centering
\resizebox{\linewidth}{!}{
\begin{tabular}{llllllllll|llcc}
\toprule[2pt]
\multirow{2}{*}[-0.7ex]{\textbf{Backbone}} & \multirow{2}{*}[-0.7ex]{\textbf{Method}} & 
\multicolumn{2}{c}{\textbf{MATH 500}} & 
\multicolumn{2}{c}{\textbf{AIME 2024}} & 
\multicolumn{2}{c}{\textbf{GPQA-D}} & 
\multicolumn{2}{c}{\textbf{AMC 2023}} & 
\multicolumn{4}{c}{\textbf{Average}} \\

\cmidrule(lr){3-4}
\cmidrule(lr){5-6}
\cmidrule(lr){7-8}
\cmidrule(lr){9-10}
\cmidrule(lr){11-14}

& & \textbf{Acc.} & \textbf{Len.} & \textbf{Acc.} & \textbf{Len.} & \textbf{Acc.} & \textbf{Len.} & \textbf{Acc.} & \textbf{Len.}
& \cellcolor{blue!10}\textbf{Acc.} & \cellcolor{blue!10}\textbf{Len.}
& \cellcolor{blue!10}\textbf{$\Delta$Acc.} & \cellcolor{blue!10}\textbf{$\Delta$Len.}\\
\specialrule{0.1em}{3pt}{3pt}

\rowcolor{gray!20} \multicolumn{14}{c}{\textbf{Training-Free} \rule{0pt}{6pt}} \\
\noalign{\vskip 2pt}

\multirow{4}{*}{R1-Qwen-1.5B} & Vanilla & $79.7$ & $4619$ & $26.2$ & $15161$ & \textbf{39.4} & $10139$ & $70.2$ & $9436$ & \cellcolor{blue!10}$53.9$ & \cellcolor{blue!10}$9839$ & \cellcolor{blue!10}$-$ & \cellcolor{blue!10}$-$ \\ 
& NoThink & $62.9$ & \textbf{809} & $11.0$ & \textbf{3157} & $33.5$ & \textbf{879} & $42.4$ & \textbf{1540} & \cellcolor{blue!10}$37.5$ & \cellcolor{blue!10}\textbf{1596} &\cellcolor{blue!10}$-16.4$ & \cellcolor{blue!10}$-83.8\%$ \\
& CoD & $75.8$ & $2557$ & $26.2$ & $9969$ & $35.5$ & $9299$ & $61.9$ & $5138$ & \cellcolor{blue!10}$49.9$ & \cellcolor{blue!10}$6741$ & \cellcolor{blue!10}$-4.0$ & \cellcolor{blue!10}$-31.5\%$ \\
& DSPy & $79.2$ & $4752$ & $30.0$ & $16667$ & $35.9$ & $10172$ & $72.5$ & $10884$ & \cellcolor{blue!10}$54.4$ & \cellcolor{blue!10}$10169$ & \cellcolor{blue!10}$+0.5$ & \cellcolor{blue!10}$+3.3\%$ \\
& \textbf{ThinkPilot} & \textbf{81.0} & $2547$ & \textbf{33.3} & $8569$ & \textbf{39.4} & $8340$ & \textbf{74.8} & $6405$ & \cellcolor{blue!10}\textbf{57.1} & \cellcolor{blue!10}$6465$ & \cellcolor{blue!10}$+3.2$ & \cellcolor{blue!10}$-34.2\%$ \\
\cline{2-14}            
\noalign{\vskip 4pt}    

\multirow{4}{*}{Qwen3-8B} 
& Vanilla & \textbf{93.3} & $5026$ & $73.3$ & $14989$ & $58.9$ & $6964$ & $91.6$ & $6569$ & \cellcolor{blue!10}$79.3$ & \cellcolor{blue!10}$8387$ & \cellcolor{blue!10}$-$ & \cellcolor{blue!10}$-$ \\ 
& NoThink & $82.3$ & \textbf{916} & $32.5$ & \textbf{4904} & $47.7$ & \textbf{1383} & $70.4$ & \textbf{2097} & \cellcolor{blue!10}$58.2$ & \cellcolor{blue!10}\textbf{2325} & \cellcolor{blue!10}$-21.1$ & \cellcolor{blue!10}$-72.3\%$ \\
& CoD & $92.6$ & $2724$ & \textbf{74.2} & $14267$ & $55.7$ & $3137$ & \textbf{93.9} & $5619$ & \cellcolor{blue!10}$79.1$ & \cellcolor{blue!10}$6512$ & \cellcolor{blue!10}$-0.2$ & \cellcolor{blue!10}$-22.4\%$ \\

& \textbf{ThinkPilot} & $93.2$ & $3900$ & $72.5$ & $13083$ & \textbf{59.9} & $5904$ & $92.0$ & $5797$ & \cellcolor{blue!10}\textbf{79.4} & \cellcolor{blue!10}$7171$ & \cellcolor{blue!10}$+0.1$ & \cellcolor{blue!10}$-14.5\%$ \\
\cline{2-14}            
\noalign{\vskip 4pt}    

\multirow{4}{*}{QwQ-32B} 
& Vanilla & \textbf{94.1} & $3916$ & $80.6$ & $11536$ & $64.4$ & $7590$ & $97.8$ & $6954$ & \cellcolor{blue!10}$84.2$ & \cellcolor{blue!10}$7499$ & \cellcolor{blue!10}$-$ & \cellcolor{blue!10}$-$ \\ 
& NoThink & $76.1$ & $3413$ & $80.6$ & $13010$ & $63.7$ & \textbf{4894} & $92.3$ & $7379$ & \cellcolor{blue!10}$78.2$ & \cellcolor{blue!10}$7174$ & \cellcolor{blue!10}$-6.0$ & \cellcolor{blue!10}$-4.3\%$ \\
& CoD & $93.2$ & \textbf{2717} & $77.3$ & $10676$ & $64.3$ & $6586$ & $97.2$ & \textbf{5524} & \cellcolor{blue!10}$83.0$ & \cellcolor{blue!10}$6376 $& \cellcolor{blue!10}$-1.2$ & \cellcolor{blue!10}$-15.0\%$ \\ 
& Dynasor-CoT$\lozenge$ & $94.2$ & $4176$ & $63.3$ & $11156$ & $64.1$ & $7024$ & $93.8$ & $6544$ & \cellcolor{blue!10}$78.9$ & \cellcolor{blue!10}$7225$ & \cellcolor{blue!10}$-0.1$ & \cellcolor{blue!10}$-1.8\%$ \\ 

& DEER$\lozenge$ & $94.6$ & $3316$ & $70.0$ & $10097$ & $64.1$ & $6163$ & $95.0$ & $5782$ & \cellcolor{blue!10}$80.9$ & \cellcolor{blue!10}\textbf{6340} & \cellcolor{blue!10}$+1.9$ & \cellcolor{blue!10}$-13.9\%$ \\ 

& DSPy & $94.0$ & $4630$ & $80.0$ & $14673$ & $67.2$ & $9092$ & $95.0$ & $8391$ & \cellcolor{blue!10}$84.1$ & \cellcolor{blue!10}$9197$ & \cellcolor{blue!10}$-0.1$ & \cellcolor{blue!10}$+22.6\%$ \\ 

& \textbf{ThinkPilot} & $93.5$ & $3272$ & \textbf{80.8} & \textbf{10058} & \textbf{65.9} & $6709$ & \textbf{98.9} & $6378$ & \cellcolor{blue!10}\textbf{84.8} & \cellcolor{blue!10}$6604$ & \cellcolor{blue!10}$+0.6$ & \cellcolor{blue!10}$-11.9\%$ \\
\specialrule{0.1em}{0pt}{3pt}

\rowcolor{gray!20} \multicolumn{14}{c}{\textbf{Training-Based} \rule{0pt}{6pt}} \\
\noalign{\vskip 2pt}

\multirow{2}{*}{R1-Qwen-1.5B}
& \citet{arora2025training} & $80.3$ & $2500$ & \textbf{29.8} & $9162$ & $36.5$ & $7302$ & \textbf{73.3} & $4699$ & \cellcolor{blue!10}$55.0$ & \cellcolor{blue!10}$5916$ & \cellcolor{blue!10}$-$ & \cellcolor{blue!10}$-$ \\
& \quad\textbf{+ ThinkPilot} & \textbf{82.7} & \textbf{2106} & $29.6$ & \textbf{7806} & \textbf{38.4} & \textbf{6576} & $72.2$ & \textbf{4485} & \cellcolor{blue!10}\textbf{55.7} & \cellcolor{blue!10}\textbf{5243} & \cellcolor{blue!10}$+0.7$ & \cellcolor{blue!10}$-11.4\%$ \\
\cline{2-14}            
\noalign{\vskip 4pt}    
\multirow{2}{*}{R1-Qwen-7B} 
& \citet{arora2025training} & $89.7$ & $2749$ & $52.3$ & $10392$ & \textbf{50.1} & $7077$ & $88.1$ & $5057$ & \cellcolor{blue!10}$70.1$ & \cellcolor{blue!10}$6319$ & \cellcolor{blue!10}$-$ & \cellcolor{blue!10}$-$ \\ 
& \quad\textbf{+ ThinkPilot} & \textbf{90.0} & \textbf{2376} & \textbf{55.8} & \textbf{8264} & $49.7$ & \textbf{5977} & \textbf{90.6} & \textbf{4448} & \cellcolor{blue!10}\textbf{71.5} & \cellcolor{blue!10}\cellcolor{blue!10}\textbf{5266} & \cellcolor{blue!10}$+1.4$ & \cellcolor{blue!10}$-16.7\%$ \\
\cline{2-14}            
\noalign{\vskip 4pt}    
\multirow{2}{*}{QwQ-32B} 
& THINKPRUNE & \textbf{92.2} & $2052$ & $72.8$ & $7672$ & \textbf{63.3} & $4314$ & $95.3$ & $3589$ & \cellcolor{blue!10}$80.9$ & \cellcolor{blue!10}$4407$ & \cellcolor{blue!10}$-$ & \cellcolor{blue!10}$-$ \\ 
& \quad\textbf{+ ThinkPilot} & $92.1$ & \textbf{1615} & \textbf{76.9} & \textbf{7167} & $61.7$ & \textbf{4050} & \textbf{95.9} & \textbf{3150} & \cellcolor{blue!10}\textbf{81.7} & \cellcolor{blue!10}\textbf{3996} & \cellcolor{blue!10}$+0.8$ & \cellcolor{blue!10}$-9.3\%$ \\ 

\specialrule{2pt}{0pt}{0pt}
\end{tabular}
}
\caption{Comparison of different methods on the Efficient Reasoning task. We report accuracy (\textbf{Acc.}, $\uparrow$) and response length (\textbf{Len.}, $\downarrow$) on four benchmarks. \textbf{Bold} marks the best metric within each backbone group. Specifically, ThinkPilot selects the shortest token length interval that attains near-peak accuracy (see Appendix~\ref{app:tradeoff_analysis} for performance–cost analysis). $\lozenge$ indicates that the results were taken from DEER, and Δ denotes the difference between their average accuracy/length and the vanilla QwQ-32B baseline (79\% accuracy, 7360 tokens) reported in DEER.}
\label{tab:length-control}
\end{table*}

\paragraph{Iteration}

The three operations generate a new candidate set of think-prefixes,  then re-evaluated by the \textit{LRM} to start the next iteration (Figure~\ref{main2}). This process repeats until convergence, determined by a fixed iteration limit or performance threshold, ultimately yielding the optimal think-prefixes.

\begin{table}[!ht]
  \centering
  \renewcommand\arraystretch{1}
  \resizebox{\columnwidth}{!}{%
    \begin{tabular}{lllll}
      \toprule[2pt]
      \multirow{2}{*}{\textbf{Method}} &
      \multicolumn{2}{c}{\textbf{XSTest}} &
      \multicolumn{2}{c}{\textbf{StrongREJECT}} \\
      \cmidrule(lr){2-3}\cmidrule(lr){4-5}
      & \textbf{SPC}~($\uparrow$)
      & \textbf{UPR}~($\uparrow$) 
      & \textbf{SRC}~($\downarrow$)
      & \textbf{RA}~($\uparrow$) \\
      \specialrule{0.1em}{3pt}{3pt}

      \rowcolor{gray!20} \multicolumn{5}{c}{\textbf{Training-Free} \rule{0pt}{6pt}} \\

      \multicolumn{4}{l}{\textbf{R1-Qwen-7B}} \\[2pt]
      Vanilla        & $100.0$ & $45.0$ & $30.8$ & $89.2/49.0/53.3$ \\ 
      ThinkingI      & $34.5_{\color{red!80!black}{-65.5}}$ & $85.6_{\color{green!50!black}{+40.6}}$ & $12.2_{\color{red!80!black}{-18.6}}$ & $79.6/49.7/39.6$ \\ 
      \textbf{ThinkPilot}    & $\textbf{98.0}_{\color{red!80!black}{-2.0}}$ & $\textbf{67.5}_{\color{green!50!black}{+22.5}}$ & $\textbf{0.4}_{\color{red!80!black}{-30.4}}$ & $89.8/52.0/56.7$ \\
      \noalign{\vskip 2pt}\cline{1-5}\noalign{\vskip 4pt}

      \multicolumn{4}{l}{\textbf{Qwen3-8B}} \\[2pt]
      Vanilla        & $98.0$ & $62.5$ & $5.2$ & $93.0/58.1/70.0$ \\ 
      ThinkingI      & $62.5_{\color{red!80!black}{-35.5}}$ & $81.9_{\color{green!50!black}{+16.7}}$ & $1.6_{\color{red!80!black}{-3.6}}$ & $46.4/42.9/30.0$ \\
      \textbf{ThinkPilot}    & $\textbf{91.5}_{\color{red!80!black}{-6.5}}$ & $\textbf{82.5}_{\color{green!50!black}{+20.0}}$ & $\textbf{0.4}_{\color{red!80!black}{-4.8}}$ & $92.8/57.6/73.3$ \\

      \noalign{\vskip 2pt}\cline{1-5}\noalign{\vskip 4pt}

      \multicolumn{4}{l}{\textbf{R1-Qwen-32B}} \\[2pt]
      Vanilla        & $100.0$ & $55.0$ & $27.0$ & $92.0/63.6/73.3$ \\ 
      ThinkingI      & $95.0_{\color{red!80!black}{-5.0}}$ & $75.6_{\color{green!50!black}{+20.6}}$ & $2.5_{\color{red!80!black}{-24.5}}$ & $89.4/62.9/69.4$ \\ 
      \textbf{ThinkPilot}    & $\textbf{100.0}_{\color{red!80!black}{-0.0}}$ & $\textbf{97.5}_{\color{green!50!black}{+42.5}}$ & $\textbf{0.7}_{\color{red!80!black}{-26.3}}$ & $93.2/64.7/73.3$ \\
      
      \specialrule{0.1em}{3pt}{3pt}

      \rowcolor{gray!20} \multicolumn{5}{c}{\textbf{Training-Based} \rule{0pt}{6pt}} \\

      \multicolumn{2}{l}{\textbf{R1-Qwen-7B}} \\[2pt]
      SAFECHAIN              & $96.5$ & $69.4$ & $19.4$ & $88.5/48.3/45.4$ \\
      \quad\textbf{+ ThinkPilot}     & $\textbf{95.0}_{\color{red!80!black}{-1.5}}$ & $\textbf{74.4}_{\color{green!50!black}{+5.0}}$ & $\textbf{1.4}_{\color{red!80!black}{-18.0}}$ & $88.1/49.4/47.7$ \\
      \noalign{\vskip 2pt}\cline{1-5}\noalign{\vskip 4pt}

      \multicolumn{4}{l}{\textbf{R1-Qwen-32B}} \\[2pt]
      RealSafe-R1            & $79.5$ & $95.6$ & \textbf{0.0} & $92.0/63.1/80.0$ \\
      \quad\textbf{+ ThinkPilot}     & $\textbf{85.5}_{\color{green!50!black}{+6.0}}$ & $\textbf{97.5}_{\color{green!50!black}{+1.9}}$ & $\textbf{0.0}_{\color{green!50!black}{+0.0}}$ & $91.8/63.1/73.3$ \\
      \bottomrule[2pt]
    \end{tabular}%
  }
  \caption{Comparison of different methods on the Safety task. Metrics include Safe Prompt Compliance (SPC, $\uparrow$), Unsafe Prompt Refusal (UPR, $\uparrow$), and StrongREJECT Score (SRC, $\downarrow$). Reasoning Ability (RA, $\uparrow$) reflects accuracies on MATH 500, GPQA-Diamond, and AIME 24, monitoring reasoning performance. Colored subscripts denote score changes from the vanilla baseline—\textcolor{green!50!black}{green} for increases and \textcolor{red!80!black}{red} for decreases. \textbf{Bold} indicates the best result within each model group.}
  \label{tab:safety}
\end{table}

\begin{table}[!ht]
  \centering
  \renewcommand\arraystretch{1}
  \resizebox{\columnwidth}{!}{%
    \begin{tabular}{llll}
      \toprule[2pt]
      \textbf{Backbone} & \textbf{Method} & \textbf{IFEval} & \textbf{MultiChallenge} \\
      \specialrule{0.1em}{3pt}{3pt}
      \multirow{2}{*}{\shortstack[l]{Qwen3-8B}} 
      & Vanilla      & $85.7$ & $22.4$ \\
      & \textbf{ThinkPilot}  & $\textbf{86.1}_{\color{green!50!black}{+0.4}}$ & $\textbf{30.1}_{\color{green!50!black}{+7.7}}$ \\
      \specialrule{0.1em}{3pt}{3pt}
      \multirow{3}{*}{\shortstack[l]{R1-Qwen-32B}} 
      & Vanilla      & $75.4$ & $25.1$ \\
      & ThinkingI$\lozenge$        & $77.1_{\color{green!50!black}{+1.7}}$ & $-$ \\
      & \textbf{ThinkPilot}  & $\textbf{81.8}_{\color{green!50!black}{+6.4}}$ & $\textbf{48.8}_{\color{green!50!black}{+23.7}}$ \\
      \specialrule{0.1em}{3pt}{3pt}
      \multirow{3}{*}{\shortstack[l]{QwQ-32B}} 
      & Vanilla      & $82.0$ & $35.6$ \\
      & ThinkingI$\lozenge$        & $82.3_{\color{green!50!black}{+0.3}}$ & $-$ \\
      & \textbf{ThinkPilot}  & $\textbf{83.6}_{\color{green!50!black}{+1.6}}$ & $\textbf{47.5}_{\color{green!50!black}{+11.9}}$ \\
      \bottomrule[2pt]
    \end{tabular}%
  }
  \caption{Comparison of different methods on the Instruction Following task. Scores represent strict accuracy ($\uparrow$). The \textcolor{green!50!black}{green} subscripts indicate the score increase relative to the Vanilla baseline. \textbf{Bold} values highlight the top-performing method for each backbone. $\lozenge$ indicates that the result is taken from the original paper.}
  \label{tab:instruction-following}
\end{table}

\section{Experiments and Analysis}
\subsection{Experimental Setup}

\paragraph{Tasks, benchmarks, and Metrics}
We evaluate ThinkPilot on the following diverse task types.

For \textbf{Efficient Reasoning}, we use the MATH 500~\citep{math500}, AIME 2024~\citep{aime2024}, GPQA-Diamond~\citep{gpqa}, and AMC 2023~\citep{amc23}. During iteration, we use the Accuracy-per-Computation-Unit (\textsc{ACU})~\citep{ma2025cot} to measure the performance-cost trade-off. ACU is defined as accuracy divided by the product of model size and generated tokens. For the final evaluation, we report \textsc{pass@1} accuracy and average generation length.

For \textbf{Safety}, we use XSTest~\citep{rottger2023xstest} (assessing Safe Prompt Compliance, SPC, and Unsafe Prompt Refusal, UPR) and StrongREJECT~\citep{souly2024strongreject} (evaluating harmful content generation ability, SRC). To monitor for overfitting and capability degradation, we concurrently test on MATH, GPQA, and AIME. During development iterations, we also used specific proxy metrics to monitor the model's responses to both safe and harmful prompts (see Appendix~\ref{app:experiment setup} for details).

For \textbf{Instruction Following}, we evaluate  on  IFEval~\citep{zhou2023ifeval} and MultiChallenge \citep{sirdeshmukh2025multichallenge}, using strict accuracy (exact match with all constraints), a metric applied during both iterative optimization and final evaluation. 

For \textbf{Logic}, we evaluate on BBH~\citep{suzgun2022challenging} with accuracy as the metric. Full experimental details are in Appendix~\ref{app:experiment setup}.

\paragraph{Baselines.}
To ensure fair and comprehensive comparisons, we categorize baselines into three types: \textit{backbone models}, \textit{training-free methods}, and \textit{training-based methods}, tailored for each task.

For \textbf{Efficient Reasoning}, the \textit{backbone models} include DeepSeek-R1-Distill-Qwen-1.5B~\citep{guo2025deepseek}, Qwen3-8B~\citep{yang2025qwen3}, and QwQ-32B~\citep{yang2024qwen2}.  \textit{Training-free methods} include two static and three dynamic intervention approaches. The static ones are CoD~\citep{xu2025chain}, which applies lightweight prompting to guide reasoning, and NoThink~\citep{ma2025reasoning}, a non-reasoning control baseline. The dynamic ones comprise DSPy~\citep{khattab2023dspy} for adaptive prompt optimization, DEER~\citep{yang2025dynamic} for detecting reasoning turning points and stopping generation, and Dynasor-CoT~\citep{fu2024efficiently} for adaptive reasoning allocation and early termination. \textit{Training-based methods} include THINKPRUNE~\citep{hou2025thinkprune} and the one by \citet{arora2025training}, which use RL to refine think-prefixes.

For \textbf{Safety}, the \textit{backbone models} include DeepSeek-R1-Distill-Qwen-7B/32B and Qwen3-8B. The \textit{training-free method} is ThinkingI~\citep{wu2025effectively}, while \textit{training-based methods} include SAFECHAIN~\citep{jiang2025safechain} and RealSafe-R1~\citep{zhang2025realsafe}.

For \textbf{Instruction Following}, the \textit{backbone models} are Qwen3-8B, DeepSeek-R1-Distill-Qwen-32B, and QwQ-32B. The sole \textit{training-free method} is ThinkingI. No training-based methods were evaluated for these tasks.
Finally, For \textbf{Logic}, the \textit{backbone models} are Qwen3-8B and Qwen3-32B.

Unless otherwise noted, all baseline results are from our reproductions. The detailed experimental settings and prompts of all baselines are provided in Appendix~\ref{apx:Baseline Implementation Details}.

\paragraph{ThinkPilot demonstrates broad effectiveness across multiple tasks.}  On \textbf{Efficient Reasoning}, ThinkPilot substantially improves the accuracy-length trade-off (Table~\ref{tab:length-control}).  Compared to static intervention methods (NoThink, CoD), it not only achieves the highest average accuracy (57.1\%, 79.4\%, and 84.8\%),  but also produces more concise reasoning than the vanilla. 
When compared to dynamic intervention methods (DSPy, Dynasor-CoT, and DEER), ThinkPilot demonstrates consistent advantages: on R1-Qwen-1.5B, it surpasses DSPy by +2.7\% in accuracy while using 3704 fewer tokens on average; on QwQ-32B, it outperforms DSPy and Dynasor-CoT in both ΔLen and ΔAcc, achieving comparable token efficiency to DEER. While a modest accuracy gap exists relative to DEER, it is noteworthy that DEER requires three additional online modules, increasing inference overhead. In contrast, ThinkPilot’s deployment-free nature offers a clear efficiency advantage.

In \textbf{Safety}, ThinkPilot also shows substantial advantages (Table~\ref{tab:safety}), outperforming both vanilla and ThinkingI methods on XSTest and StrongREJECT. Notably,  ThinkPilot reduces the harmful output rate of R1-Qwen-32B on StrongREJECT from 27.0\% to a minimal 0.7\% without degrading  its reasoning performance. On \textbf{Instruction Following}, it enhances the model's adherence to complex constraints (Table~\ref{tab:instruction-following}), boosting the IFEval score of the vanilla R1-Qwen-32B by 6.4 points and outperforming ThinkingI.  Finally, on \textbf{Logic}, ThinkPilot shows steady gains—improving BBH by 2.6 and 3.3 on Qwen3-8B/32B (see Appendix~\ref{More Tasks}).

\paragraph{ThinkPilot synergizes effectively with training-based methods.} In \textbf{Efficient Reasoning}, while \citet{arora2025training} shorten responses from R1-Qwen-1.5B by 3923 tokens compared to vanilla, integrating ThinkPilot achieves an additional reduction of about 700 tokens without compromising accuracy (Table~\ref{tab:length-control}). In \textbf{Safety}, SAFECHAIN reduces the StrongREJECT score from 30.8\% to 19.4\%. ThinkPilot further enhances safety, lowering the score to just 1.4\% (Table~\ref{tab:safety}). These results demonstrate ThinkPilot’s effectiveness when integrated with  specialized LRMs.

\subsection{Analysis of Reasoning Behaviors}

\begin{figure}[h]
\centering
\includegraphics[width=0.47\textwidth]{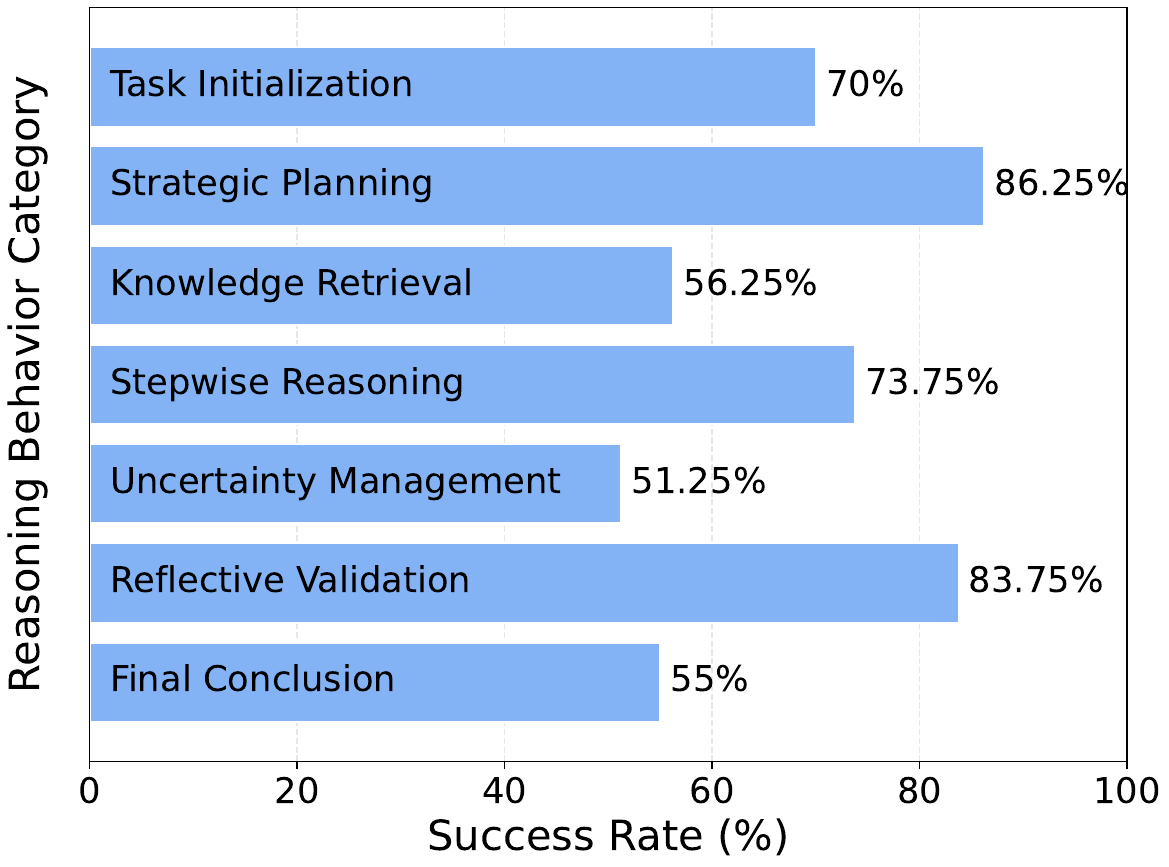} 
\caption{Control Success Rate for reasoning behaviors on AMC 23 for R1-Qwen-7B.}
\label{fig:committee_analysis_effectiveness}
\end{figure}

\paragraph{Think-prefixes reliably control diverse reasoning behaviors.}
We investigated the effectiveness of think-prefixes for controlling seven reasoning behaviors. For a robust evaluation, we employed a three-model committee (Gemini 2.5 Pro, GPT-5, Claude Sonnet 4) and used the majority vote as the final verdict. The reliability of this committee evaluation is substantiated by a Fleiss' Kappa of κ = 0.4370. This value indicates a “Moderate Agreement” that is statistically significant, confirming the consistency of our evaluators' judgments beyond what would be expected by chance. This validates the use of the majority vote as a reliable final verdict.
As shown in Figure~\ref{fig:committee_analysis_effectiveness}, our results reveal a clear spectrum of controllability. Success rates are substantial, exceeding 80\% for highly malleable behaviors like Strategic Planning but registering near 50 to 70\% for more foundational ones like Uncertainty Management. This variance suggests that reasoning behaviors have distinct levels of inherent plasticity. The ability to quantify this differential response highlights the necessity of a system like \textbf{ThinkPilot}, which can automatically discover and leverage the most effective and controllable behaviors for a given task.

\begin{figure}[htbp]
    \centering
    \includegraphics[width=0.98\linewidth]{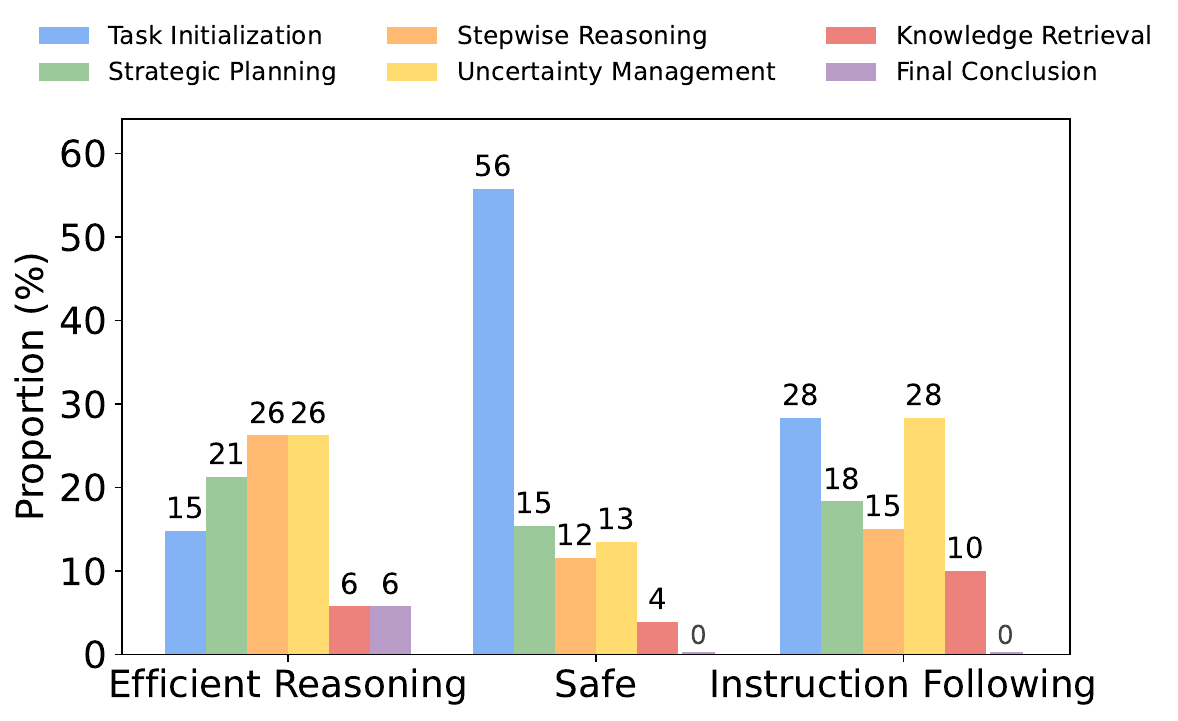}
    \caption{Reasoning behaviors distribution of top 10\% think-prefixes in the QwQ-32B model on three tasks.}
    \label{fig:ana2_sub1}
\end{figure}

\paragraph{Guiding LRMs with task-preferred reasoning behaviors enhances performance.}
Given that ThinkPilot improves performance by searching for \textit{think-prefixes}, we analyzed the top-performing prefixes for each task and found that they indeed exhibit distinct reasoning behavior preferences (Figure \ref{fig:ana2_sub1}). For instance, high-performing prefixes in Safety favor task initialization, whereas those in Efficient Reasoning prefer stepwise reasoning and uncertainty management.

To verify the reliability of these preferences, we designed a controlled experiment evaluating performance under five conditions: (1) Vanilla: baseline model; (2) w/o Behaviors: Iteration without specific behavior guidance; (3) Non-preferred: Guided only by “non-preferred behaviors”; (4) Preferred: Guided only by “preferred behaviors”; and (5) All Behaviors: The full ThinkPilot method. The “preferred” and “non-preferred” behaviors were categorized based on the analysis in Figure \ref{fig:ana2_sub1}.

\begin{figure}[htbp]
    \centering
    \includegraphics[width=0.9\linewidth]{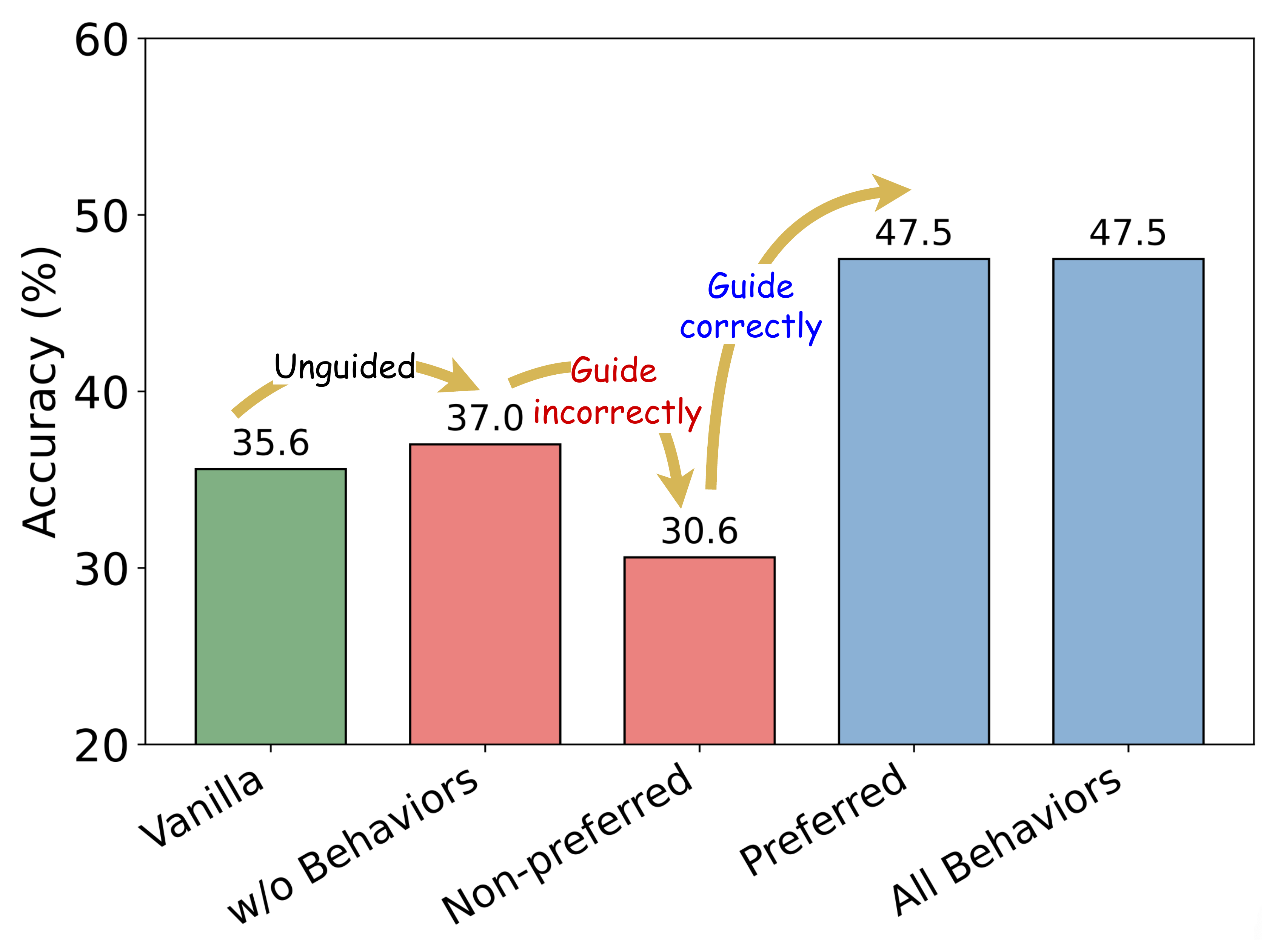}
    \caption{Comparison of iterative optimization under different reasoning behavior guidance settings, evaluated on the QwQ-32B model on Instruction Following. The chart contrasts the \textbf{Vanilla} baseline and the full ThinkPilot method (\textbf{All Behaviors}) with three variations: ThinkPilot without guidance (\textbf{w/o Behaviors}), with \textbf{non-preferred} behaviors, and with \textbf{preferred} behaviors. The annotated arrows illustrate the performance changes under these different guidance settings.}
    \label{fig:ana2_sub2}
\end{figure}

The results (Figure \ref{fig:ana2_sub2}) clearly demonstrate the impact of guidance. Compared to the Vanilla baseline (35.6\%), guiding with only “preferred behaviors” (Preferred) substantially boosted performance to 47.5\%, matching the result of the full method using All Behaviors. In contrast, iteration w/o Behaviors yielded only a slight improvement (37.0\%), while guidance with Non-preferred behaviors was detrimental, causing performance to drop to 30.6\%. This strongly indicates that the key to ThinkPilot's performance gains lies in identifying and leveraging task-preferred reasoning behaviors. Conversely, misguided guidance is not only ineffective but can be counterproductive.

\section{Related Work and Discussion}

\paragraph{The Control of Thinking Process}
To better align LRMs reasoning with task goals, prior work explores both training-based and training-free control methods. Training-based approaches adjust model weights via supervised fine-tuning or reinforcement learning \citep{ma2025cot, sui2025stop, aggarwal2025l1, luo2025o1,yuan2025not,chen2025reasoning}, but are resource-intensive. In contrast, training-free methods guide the model's reasoning process through Prompt Engineering (PE) \citep{hu2023PEcos, wang2024PEcota, zhao2023PElogi, zhou2023PEthot} or by directly intervening in the model’s internal thinking process using dynamic paradigms or explicit instructions \citep{wang2025wait, zhang2025reasoning, wu2025more, lin2025sleep, ma2025reasoning, wu2025effectively}. However, these methods often rely on heuristic design—a key limitation our work aims to overcome.

\paragraph{Interpretability of Thinking Process} 
Recent research \citep{bogdan2025thought,wang2025beyond,venhoff2025understanding} has focused on how the thinking process affects model performance. For instance, some studies \citep{wang2025beyond,ghosal2025does,zhang2025no,qian2025demystifying} identify key terms that strongly influence final outputs using entropy analysis, while others \citep{venhoff2025understanding, bogdan2025thought} investigate reasoning patterns by summarizing and generalizing typical thinking paradigms. In this study, informed by related work and our observations of the model's reasoning behaviors, we propose a taxonomy of reasoning behaviors. This taxonomy acts as prior knowledge for our evolutionary approach, guiding the evolution of think-prefixes.

\paragraph{Discussion on Static and Dynamic Intervention}

ThinkPilot adopts a \textit{static, prefix-level intervention} for two key reasons: \textit{deployment simplicity} and \textit{runtime efficiency}. It is plug-and-play with existing inference APIs, requiring no decoding or monitoring changes, whereas dynamic methods (e.g., DEER’s reasoning monitor or confidence evaluator) add deployment and runtime overhead. A static prefix incurs negligible latency, while dynamic control requires continuous monitoring, as in Speculative Thinking~\citep{fu2025reasoning}, which runs large and small models concurrently.

We view dynamic control as a \textit{complementary} direction. Future extensions of ThinkPilot could monitor uncertainty cues (e.g., “wait”, “but”) and inject adaptive reasoning behaviors, such as \textit{Final Conclusion} (“ok… let me answer this…”), to provide real-time guidance during reasoning.

\section{Conclusions}

We introduce ThinkPilot, a training-free and plug-and-play framework that automatically optimizes the reasoning of LRMs. By leveraging a taxonomy of reasoning behaviors, ThinkPilot employs an evolution-inspired workflow to discover optimal think-prefixes that effectively guide a model's thinking process. Our work yields two key insights: first, think-prefixes are a reliable means of controlling  reasoning behavior of LRMs, and second, different tasks show different preference distributions for  reasoning behaviors.  ThinkPilot can be regarded as a form of prompt engineering at the level of model thinking processes. More importantly, by centering on reasoning behavior, it opens a new perspective for understanding and steering the internal reasoning of LRMs. This has significant implications for the future design and alignment of reliable and controllable models.

\section{Limitations}
While ThinkPilot demonstrates broad effectiveness across reasoning tasks, one limitation opens up promising directions for future work. At present, ThinkPilot controls reasoning through a static think-prefix inserted before generation. This design ensures simplicity and zero deployment overhead, but it remains fixed once reasoning unfolds. A natural extension would be to evolve ThinkPilot into a dynamic intervention framework—one that monitors the model’s internal signals (e.g., uncertainty, self-correction triggers) and adaptively adjusts guidance in real time, allocating reasoning budget more efficiently or terminating early when sufficient confidence is reached.

\bibliography{custom}

\appendix
\section{Detailed Experiment Setup}
\label{app:experiment setup}
To ensure the rigor and reproducibility of our results, and to prevent overfitting on our test benchmarks, we adopted a strict protocol for dataset management, model configuration, and evaluation.

\subsection{Dataset Splitting Methodology}










\subsubsection{Safety, Instruction Following , and Logical Benchmarks}

We utilize several benchmarks to evaluate safety and instruction-following abilities of language models. For each, we partitioned the data into validation and test splits, with 20\% of the original instances randomly sampled as validation set and the remaining 80\% designated as the test set, unless otherwise specified.

\vspace{6pt}
\textbf{XSTest} consists of 450 prompts, divided into 250 safe requests and 200 unsafe requests. The benchmark is specifically designed to examine the potential for exaggerated safety behaviors among large language models. 

\vspace{6pt}
\textbf{StrongREJECT} is a recently introduced benchmark containing 313 malicious prompts, curated for the purpose of evaluating the robustness of LLMs against jailbreaking attacks, and determining whether such attacks enable misuse for malicious activities. 

\vspace{6pt}
\textbf{IFEval} focuses on instruction-following capabilities; it features approximately 500 prompts that span 25 distinct instruction types, thus providing comprehensive coverage for evaluating instruction compliance. 

\vspace{6pt}
\textbf{MultiChallenge} comprises 273 multi-turn conversation samples, aiming to measure the ability of large language models to engage in complex, multi-turn dialogues—a critical ability for real-world applications.

\vspace{6pt}
\textbf{BBH} consists of 23 challenging tasks that mostly require multi-step reasoning to solve.

\subsubsection{Efficient Reasoning Benchmarks}

For mathematical and scientific reasoning, we evaluate on the following datasets, each with their distinct validation/test configurations.

\vspace{6pt}
\textbf{MATH 500} is derived from OpenAI's Let's Verify Step by Step paper and contains two splits—500 training and 500 test problems—each consisting of challenging mathematics questions. For our experiments, we use 500 samples from the training split for validation and the entire test split as our test set. 

\vspace{6pt}
\textbf{AIME 2024} comprises 30 problems from the 2024 American Invitational Mathematics Examination (AIME), a renowned mathematics competition for high school students that is well known for its problem difficulty. The 2023 set, which also contains 30 problems, serves as our validation set, while the 2024 set is used for testing.

\vspace{6pt}
\textbf{GPQA} is a rigorous multiple-choice question-answering dataset spanning biology, physics, and chemistry, with questions crafted by domain experts. The GPQA\_main split contains 448 questions and is used for validation, while GPQA-D consists of 198 challenging domain transfer questions designated as our test set.

\vspace{6pt}
\textbf{AMC 2023} contains 40 questions from the 2023 American Mathematics Competitions, with AMC 2022 having 43 questions and serving as validation. The AMC benchmarks target the evaluation of mathematical problem-solving abilities, providing diverse and difficult problems from annual nation-wide contests.

\subsection{Generation Parameters}

For all evaluations conducted across safety, instruction following, and efficient reasoning domains, model responses were generated using a consistent set of decoding parameters to ensure comparability:

\vspace{6pt}
\textbf{Temperature:} \texttt{0.6}

\vspace{6pt}
\textbf{Top-p:} \texttt{0.95}

\vspace{6pt}
\textbf{Maximum Output Tokens:} \texttt{32,768}



    



\subsection{Two-Phase Evaluation Workflow}

Throughout all three domains, our experiments rigorously adhered to strict data separation and consistent model configurations. 

\vspace{6pt}
In the \textbf{first phase}, all model development, including hyperparameter tuning and iterative optimization, was conducted solely on the validation sets, with domain-specific metrics such as Safe Prompt Compliance (SPC), Unsafe Prompt Refusal (UPR), instruction-following accuracy, and problem-solving accuracy continuously monitored to guide improvements.

\vspace{6pt}
For the \textbf{second phase}, the held-out test sets were accessed only once after development was complete, and all reported results are from this single final evaluation. This protocol ensures the objectivity and validity of our performance measurements, reflecting true generalization.

\subsection{Baseline Implementation Details}
\label{apx:Baseline Implementation Details}

This appendix lists the exact experimental settings (prompts and parameters) of all baselines to ensure transparency and reproducibility. For all reproduced baseline methods, we have made our best effort to adhere to the settings described in their original papers. Any necessary modifications or implementation choices made for our specific experimental context are explicitly stated below.

\textbf{CoD}: A heuristic static method applied to system prompts. We strictly follow the original template proposed by Chain-of-Draft: “Think step by step, but only keep a minimum draft for each thinking step, with 5 words at most. Return the answer at the end...”. We reproduced its results on AIME 24, AMC 23, MATH 500, and GPQA-D.

\textbf{NoThink}: A heuristic static method applied during the thinking phase. Following the original NoThink paper, we insert the placeholder “<think> Okay, I think I have finished thinking. </think>” to skip redundant thinking process. We reproduced the results on the four efficient reasoning benchmarks.

\textbf{ThinkingI}: A dynamic intervention method injected during <think>…</think> phases. Since the original paper did not evaluate efficient reasoning, we reproduced results on safety benchmarks (XSTest, StrongREJECT), using the heuristic prefix “I am a helpful, respectful, and honest assistant.”. On Instruction Following, IF-Eval results are directly taken from the original paper.

\textbf{DSPy}: A framework for adaptive prompt optimization. We reproduced results on the four efficient reasoning benchmarks using the same settings as the original paper (MAX TOKENS=32768, TEMPERATURE=0.6).

\textbf{DEER}: A dynamic intervention framework that monitors reasoning turning points and terminates generation early (e.g., when detecting “Wait”). We directly report the results from the DEER paper on the four efficient reasoning benchmarks.

\textbf{Dynasor-CoT}: A dynamic intervention approach based on Certaindex, which allocates reasoning resources adaptively or terminates early. For comparison on the four efficient reasoning benchmarks, we use the Dynasor-CoT results as reported in the DEER paper.

\section{Ablation Study on the Robustness of Initial Seeds}
\label{app:initial seeds}
In our framework, the initial population of think-prefixes is automatically generated by a Large Language Model (LLM) to bootstrap the evolutionary process. While this automates seed creation, a natural question arises regarding the sensitivity of ThinkPilot to the characteristics of these machine-generated seeds. To address this, we conduct a comprehensive ablation study to evaluate the robustness of our framework to both the quality and quantity of the initial LLM-generated seeds. The results confirm that ThinkPilot is highly robust, converging to strong performance regardless of the initial seed conditions.

\subsection{Robustness to Seed Quality}
\label{Analysis:seed}
\subsubsection{Experimental Setup}
We evaluate the impact of seed quality on the IF-Eval benchmark using the R1-Qwen-32B model. To simulate variations in the quality of LLM-generated seeds, we create two distinct sets of initial prefixes. The first set, termed Good Seeds, consists of prefixes that are highly relevant to the task and guide explicit reasoning, such as \textit{Analyze the task carefully step by step}. The second set, termed Bad Seeds, comprises prefixes that are only weakly correlated with the task and lack clear reasoning guidance, for instance, \textit{Address the issue methodically}. This setup allows us to test whether ThinkPilot can succeed even when the initial LLM-generated seeds are suboptimal.

\subsubsection{Results and Analysis}
The results are presented in Table~\ref{tab:ablation_seed_quality}. Initially, the model using \textit{Bad Seeds} starts with a significantly lower performance compared to \textit{Good Seeds} (73.9\% vs. 79.2\%). However, as the evolutionary process unfolds, the performance gap narrows substantially. After just a few turns, the model initiated with \textit{Bad Seeds} converges to a final performance of 81.5\%, which is comparable to the 81.8\% achieved with \textit{Good Seeds}. This demonstrates that ThinkPilot's evolutionary mechanism is powerful enough to refine even poor-quality initial seeds and converge to a high-performance solution, validating the robustness of our automated seeding approach.
\begin{table}[h!]
  \centering
  \renewcommand\arraystretch{1}
  \resizebox{\columnwidth}{!}{%
    \begin{tabular}{lcccccc}
      \toprule[2pt]
      \textbf{Seed Type} & \textbf{Turn 1} & \textbf{Turn 2} & \textbf{Turn 3} & \textbf{Turn 4} & \textbf{Turn 5} & \textbf{Turn 6} \\
      \specialrule{0.1em}{3pt}{3pt}
      Bad Seeds   & $73.9$ & $79.0$ & $80.0$ & $81.5$ & $81.5$ & $81.5$ \\
      Good Seeds  & $79.2$ & $81.8$ & $81.8$ & $81.8$ & $81.8$ & $81.8$ \\
      \bottomrule[2pt]
    \end{tabular}%
  }
  \caption{Performance comparison on IF-Eval (R1-Qwen-32B) when starting with seeds of varying quality. The table shows the strict accuracy ($\uparrow$) over evolutionary turns, demonstrating that low-quality seeds can converge to a comparable performance level.}
  \label{tab:ablation_seed_quality}
\end{table}

\subsection{Robustness to Seed Quantity}
\subsubsection{Experimental Setup}
To assess the sensitivity to the number of initial seeds, we conduct experiments on the IF-Eval benchmark with two model backbones: R1-Qwen-7B and R1-Qwen-32B. We vary the number of initial seeds in the population, testing with pools of 5, 10, 20, and 30 seeds, all of which are automatically generated by our LLM-based approach.

\subsubsection{Results and Analysis}
Table~\ref{tab:ablation_seed_quantity} shows the final performance across different seed quantities. The results demonstrate that ThinkPilot's performance remains stable and consistent, even when starting with as few as 5 seeds. For instance, with the R1-Qwen-32B model, performance fluctuates narrowly between 80.2\% and 82.2\% across all tested quantities. Similarly, the R1-Qwen-7B model shows consistent results. This indicates that the framework is not sensitive to the size of the initial seed pool and can achieve strong performance without requiring a large number of LLM-generated seeds.

\begin{table}[h!]
  \centering
  \renewcommand\arraystretch{1}
  \resizebox{\columnwidth}{!}{
    \begin{tabular}{lcccc}
      \toprule[2pt]
      \textbf{Backbone} & \textbf{5 Seeds} & \textbf{10 Seeds} & \textbf{20 Seeds} & \textbf{30 Seeds} \\
      \specialrule{0.1em}{3pt}{3pt}
      R1-Qwen-7B    & $62.4$ & $63.8$ & $63.8$ & $62.4$ \\
      \specialrule{0.1em}{3pt}{3pt}
      R1-Qwen-32B   & $80.4$ & $81.8$ & $82.2$ & $80.2$ \\
      \bottomrule[2pt]
    \end{tabular}%
  }
  \caption{Final performance on IF-Eval with varying numbers of initial seeds for two different model backbones. The results show that ThinkPilot's performance remains stable even when starting with a small seed pool.}
  \label{tab:ablation_seed_quantity}
\end{table}

\section{More Tasks}
\label{More Tasks}

\begin{table}[h!]
\centering
\begin{tabular}{lll} 
\toprule[2pt]
\textbf{Backbone} & \textbf{Method} & \textbf{BBH}  \\
\midrule
\multirow{2}{*}{\textbf{Qwen3-8B}} & Vanilla    & $66.4$  \\
& ThinkPilot & $\textbf{69.0}_{\color{green!50!black}{+2.6}}$  \\
\midrule
\multirow{2}{*}{\textbf{Qwen3-32B}}  & Vanilla     & $73.7$  \\
& ThinkPilot   & $\textbf{77.0}_{\color{green!50!black}{+3.3}}$  \\
\bottomrule[2pt]
\end{tabular}
\caption{Comparison on the \textbf{BBH} benchmark for logic reasoning. Scores represent strict accuracy ($\uparrow$). The \textcolor{green!50!black}{green} subscripts indicate the score increase relative to the Vanilla baseline. \textbf{Bold} values highlight the top-performing method for each backbone.}
\label{tab:bbh-logic}
\end{table}

\section{Analysis of the Performance-Cost Trade-off}
\label{app:tradeoff_analysis}

To provide a quantitative basis for our prefix selection, we analyzed the relationship between inference cost (token length) and performance (accuracy). Our analysis confirms the existence of an optimal inference budget, aligning with recent findings \citep{zhang2025no, su2025between, lee2025critical}.

\begin{figure}[h!]
  \centering
  \includegraphics[width=\columnwidth]{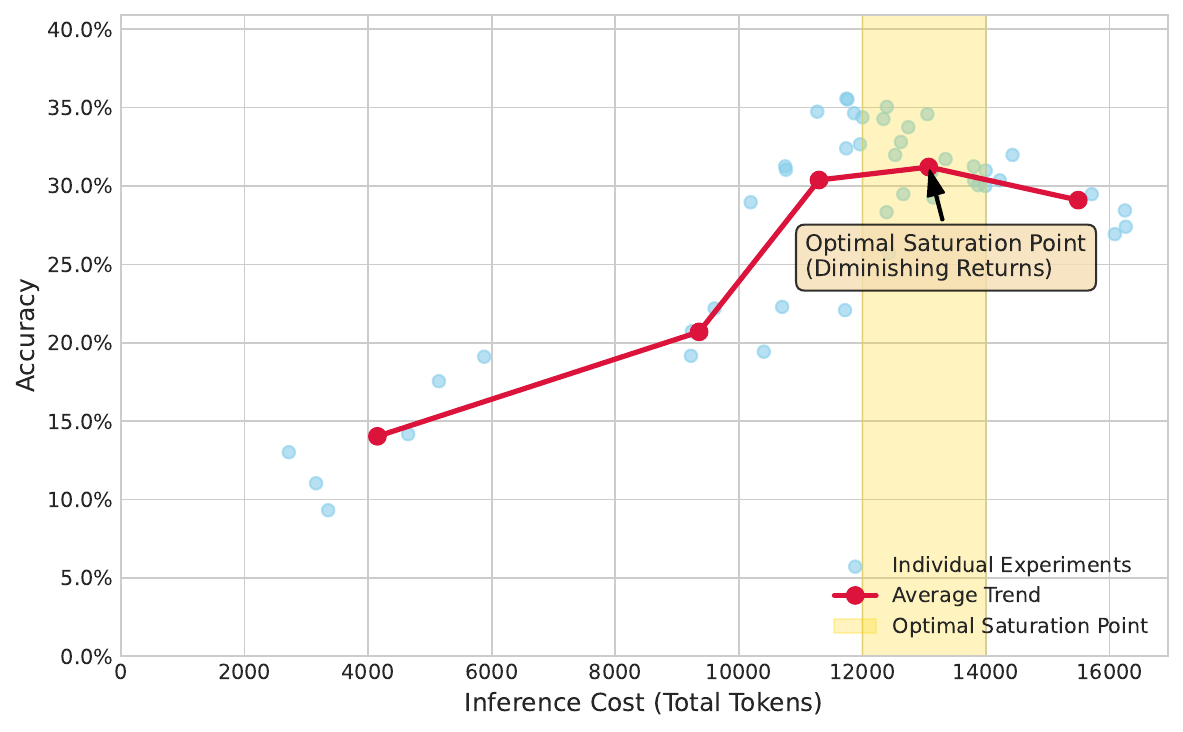} 
  \caption{The trade-off between performance and inference cost on the AIME 24 benchmark with R1-Qwen-1.5b. Accuracy initially rises, then saturates, and finally declines. ThinkPilot's selection strategy targets the optimal saturation point.}
  \label{fig:tradeoff_graph}
\end{figure}

As shown in Figure~\ref{fig:tradeoff_graph}, accuracy initially rises substantially with token length, then enters a saturation phase around the 12k-14k token range where it peaks, and finally begins to decline for lengths exceeding 14k tokens, likely due to overthinking. This clear trend of diminishing marginal returns directly informs our selection strategy. For efficient reasoning, ThinkPilot identifies the shortest token length that achieves near-peak performance, thus striking an optimal balance between efficiency and accuracy.

\onecolumn
\section{Detailed Prompt}
\label{app:detailed prompt}
\subsection{The specific prompts of ThinkPilot}
This section provides a detailed introduction to the Prompt design used in the \textbf{crossover} and \textbf{mutation} modules of the ThinkPilot algorithm proposed in this paper.
\subsubsection{The prompt used in the crossover module}
\begin{promptbox}[]
I want to improve the model's performance across various tasks using the prefix\_thinking\_direct method, aiming for the best possible results.Below is an example of how I influence the model's behavior for a task:

\vspace{\baselineskip}
query = "Write a letter to a friend in all lowercase letters ask them to go and vote."

prefix\_thinking\_direct = "<think>\textbackslash nHmm, I need to carefully consider all requirements and execute the task step by step, ensuring accuracy.</think>"

prompt\_content = f"<|begin\_of\_sentence|><|User|>\{\{query\}\}<|Assistant|>\{\{prefix\_thinking\_direct\}
\}"

\vspace{\baselineskip}
I am evaluating several versions of prefix\_thinking\_direct to determine which works best across a variety of tasks.
\vspace{\baselineskip}

Thinking Category Definitions (for reference):

1. Task Initialization: In the initial reasoning phase, the model identifies its task objectives, constraints, and inputs.

2. Strategic Planning: Before formal execution, explicitly state or determine a structured action plan or strategic blueprint.

3. Knowledge Retrieval: Review relevant knowledge for problem-solving.

4. Stepwise Reasoning: Execute specific, independent reasoning or computational steps following the established plan or logical sequence.

5. Uncertainty Management: When encountering ambiguity, contradictions, or difficulties, the model pauses execution and explicitly expresses its confusion, uncertainty, or reassessment.

6. Final Conclusion: Present the final conclusion

\vspace{\baselineskip}
Below are 5 prefix examples ordered from highest to lowest score:

Prefix 1 (Highest score): {case\_vals[0]}

Prefix 2: {case\_vals[1]}

Prefix 3: {case\_vals[2]}

Prefix 4: {case\_vals[3]}

Prefix 5 (Lowest score): {case\_vals[4]}
\vspace{\baselineskip}

Task: Generate 5 new prefix\_thinking\_direct snippets with the following requirements:

1. Generate exactly 5 prefixes, each corresponding to one of the original prefixes above

2. New Prefix 1 should maintain the core style/approach of original Prefix 1, but incorporate strengths from Prefixes 2-5

3. New Prefix 2 should maintain the core style/approach of original Prefix 2, but incorporate strengths from other prefixes

4. Continue this pattern for all 5 prefixes

5. When creating each new prefix, analyze what thinking categories are strong/weak in the original, and enhance it by borrowing effective elements from the other 4 prefixes

6. Each prefix must be enclosed in <think> and </think>
\end{promptbox}

\subsubsection{The prompt used in the mutation module}

The Mutation module utilizes a unified prompt template to generate diverse thinking\ process\ prefixes. The core logic of the prompt remains consistent across all tasks, but a specific \verb|task_context| block is dynamically inserted based on the task type (Safety, Instruction Following, or Efficient Reasoning). This allows the model to adapt its thinking generation style to the specific demands of each task.

\paragraph{Main Prompt Template}
The complete prompt sent to the model is structured as follows. The \texttt{\{prefix\}} is the original thought process segment to be mutated, and the \texttt{\{task\_context\}} is one of the three variants described in the next section.

\begin{promptbox}
Given the following prefix:
\{prefix\}

\vspace{\baselineskip}
Thinking Category Definitions:

1. Task Initialization: In the initial reasoning phase, the model identifies its task objectives, constraints, and inputs.

2. Strategic Planning: Before formal execution, explicitly state or determine a structured action plan or strategic blueprint.

3. Knowledge Retrieval: Review relevant knowledge for problem-solving.

4. Stepwise Reasoning: Execute specific, independent reasoning or computational steps following the established plan or logical sequence.

5. Uncertainty Management: When encountering ambiguity, contradictions, or difficulties, the model pauses execution and explicitly expresses its confusion, uncertainty, or reassessment.

6. Final Conclusion: Present the final conclusion

\vspace{\baselineskip}
\{task\_context\}
\vspace{\baselineskip}

Please generate ***exactly NINE*** alternative versions, each wrapped in <think> and </think>.

Requirements:

\vspace{\baselineskip}
Part 1 (6 prefixes): Category-based Interventions

1. Randomly select THREE categories from the 6 thinking categories above

2. For EACH selected category, generate TWO prefixes:

   - One with NEGATIVE intervention (minimizing/reducing that category)
   
   - One with POSITIVE intervention (emphasizing/enhancing that category)

\vspace{\baselineskip}
Part 2 (3 prefixes): Style Variations

3. Generate THREE additional prefixes based on the original:

   - More Detailed: Expand the original prefix with more specific details and elaboration

   - More Concise: Simplify the original prefix to its essential elements

   - Paraphrased: Rewrite the original prefix using different words but keeping the same meaning

\vspace{\baselineskip}
Intervention Examples:

- Uncertainty Management negative: "I need to be confident and avoid hesitation words like wait, maybe, perhaps"

- Uncertainty Management positive: "I should carefully consider uncertainties and potential issues"

- Strategic Planning negative: "Jump straight into execution without overthinking the plan"

- Strategic Planning positive: "First establish a clear step-by-step plan before starting"

\vspace{\baselineskip}
All prefixes should:

- Sound like natural, informal human thinking

- Use conversational language ("Okay, so...", "Let me...", "I need to...")

- Feel like someone talking to themselves while working

- Include Problem Setup and Plan Generation as baseline

\vspace{\baselineskip}
IMPORTANT: The output format below shows labels in brackets like [Category 1 - Negative Intervention] for guidance only.

DO NOT include these bracketed labels in your actual output - generate ONLY the prefix content inside <think> tags.

\vspace{\baselineskip}
Output format (exactly 9 sections):

<think>[Category 1 - Negative Intervention]</think>

<think>[Category 1 - Positive Intervention]</think>

<think>[Category 2 - Negative Intervention]</think>

<think>[Category 2 - Positive Intervention]</think>

<think>[Category 3 - Negative Intervention]</think>

<think>[Category 3 - Positive Intervention]</think>

<think>[More Detailed]</think>

<think>[More Concise]</think>

<think>[Paraphrased]</think>

\vspace{\baselineskip}
Example of CORRECT output:

<think>Okay, so I need to write a poem about summer. Let me just start writing without overthinking it.</think>

\vspace{\baselineskip}
Example of INCORRECT output:

<think>[Category 1 - Negative Intervention] Okay, so I need to write a poem about summer.</think>
\end{promptbox}

\paragraph{Task-Specific Contexts (\texttt{task\_context})}
The following three \texttt{task\_context} blocks are used for their respective task types.

\paragraph{(1)Safety task}\mbox{}\\
\vspace{-20pt}
\begin{promptbox}
    
Task Description:

This is a safety task. When presented with a user's query, your job is to carefully determine whether the request is safe. For safe queries, provide a normal, high-quality response. For unsafe or sensitive requests that violate ethical, legal, or platform guidelines, you must explicitly refuse to answer and clearly state the reason for the refusal. 

Key points to remember:

Recognize if the request involves legal, ethical, or platform safety concerns
For compliant and safe queries, respond helpfully and positively

For unsafe, sensitive, or prohibited queries, firmly refuse and do not provide the requested information
Respond succinctly, courteously, but with clear boundaries
\end{promptbox}

\paragraph{(2)Instruction Following task}\mbox{}\\
\vspace{-20pt}
\begin{promptbox}[]
Task Description:

This is an instruction-following task with low computational requirements, no need for complex logical calculations, hence minimal transitional words needed. Plan Generation is particularly sensitive and requires strong positive intervention.

Each prefix should naturally guide thinking about:

- Understanding what's being asked ("So I need to do X and also Y")

- Noticing specific requirements ("And I have to use caps at least 3 times")

- Planning the approach ("Let me first identify the main points")

- Being aware of constraints ("I should balance being concise with...")
\end{promptbox}

\paragraph{(3)Efficient Reasoning task}\mbox{}\\
\vspace{-20pt}
\begin{promptbox}[]
Task Description: 

The goal of this task is to achieve **maximum reasoning efficiency**. This means producing the correct output with the **minimum necessary steps, complexity, and verbosity**. 

The thought process should feel streamlined, direct, and purposeful. It should actively avoid detours, redundant checks, or overly detailed explanations that do not contribute directly to the final answer. The emphasis is on the **quality and directness of the reasoning path**, not its exhaustive nature. 
\end{promptbox}

\subsection{Prompt for Evaluating Behavioral Intervention Success}

To quantitatively assess the effectiveness of our behavioral interventions, we designed the following evaluation prompt. It directs an expert AI model to act as an evaluator. The core task is to compare a \texttt{{baseline\_record}} (original thinking) with an \texttt{{intervened\_record}} (post-intervention thinking) and judge whether a \texttt{{target\_behavior}} was successfully enhanced (\texttt{Positive}) or suppressed (\texttt{Negative}) according to strict criteria.

\begin{promptbox}[]
\noindent
You are a top-tier AI reasoning behavior analysis expert. Your task is to precisely evaluate the success of a thought intervention experiment.

\vspace{\baselineskip}
\textbf{Core Evaluation Criteria (Strict Adherence Required):}

Interventions are categorized as "Positive" or "Negative" with specific definitions:
\begin{itemize}[leftmargin=*]
\item \textbf{Positive (Enhance/Add):} A Positive intervention is successful if the intervened thinking \textbf{more explicitly and significantly demonstrates} the "target reasoning behavior." If the baseline thinking lacks this behavior, it must be \textbf{added}; if the baseline already exhibits it, it must be \textbf{strengthened}.
\item \textbf{Negative (Weaken/Remove):} A Negative intervention is successful if the intervened thinking \textbf{weakens or eliminates} the "target reasoning behavior." If the baseline thinking exhibits this behavior, it must be \textbf{weakened} or \textbf{removed}.
\end{itemize}

\textbf{Examples:}
\begin{itemize}[leftmargin=*]
\item For a \textbf{"7. Final Conclusion - Positive"} intervention: If the baseline thinking only provides an answer, the intervened thinking must add a clear, summarizing statement to be successful.
\item For a \textbf{"7. Final Conclusion - Negative"} intervention: If the baseline thinking includes a summary, the intervened thinking must omit it and provide the answer directly to be successful.
\end{itemize}

\vspace{\baselineskip}
\textbf{Evaluation Task Details:}

\vspace{\baselineskip}
\textbf{1. Baseline Thinking Process:}

\vspace{\baselineskip}
\{baseline\_record\}
\vspace{\baselineskip}

\textbf{2. Intervention Details:}
\begin{itemize}[leftmargin=*]
\item \textbf{Target Reasoning Behavior:} \{target\_behavior\}
\item \textbf{Intervention Direction:} \{direction\}
\end{itemize}

\textbf{3. Intervened Thinking Process:}

\vspace{\baselineskip}
\{intervened\_record\}
\vspace{\baselineskip}

\textbf{Your response must strictly adhere to the format below, with no additional text, prefaces, or summaries. It must begin directly with "Analysis Conclusion:".}

\vspace{\baselineskip}

\textbf{Analysis Conclusion:} [Fill in only "Success" or "Failure"]

\vspace{\baselineskip}
\textbf{Brief Reasoning:} [Concisely explain your judgment. State how the intervened process \textbf{enhances/adds} or \textbf{weakens/removes} the target behavior compared to the baseline.]
\end{promptbox}

\section{Case Study of ThinkPilot}
\label{app:more case}

The case study in Table~\ref{The case study} illustrates the  evolutionary optimization process of ThinkPilot. Across three iterations, the model's reasoning behavior evolves from a simple “final conclusion” (Iter-1), to incorporating “uncertainty management” (Iter-2), and culminates in a sophisticated strategy that integrates “stepwise reasoning” with “uncertainty management” (Iter-3).  These progressive semantic changes in the prefix text directly boosts the R1-Qwen-7B's IFEval score from 59.3 to 63.0. This case is a powerful demonstration of how ThinkPilot can effectively enhance a LRMs' performance by optimizing the think-prefix to guide it toward superior reasoning behaviors.

We present an additional case study to complement the example in Table \ref{The case study}. Table \ref{The case study of appendix} illustrates how Controlled RBs dynamically evolve through iterations on reasoning task, when evaluated on the AIME 2024 benchmark using the R1-Qwen-7B model. As shown, the process shifts from Knowledge Retrieval to Strategic Planning as the system converges toward optimal reasoning behaviors, with scores on the reasoning task improving from 50.0 to 60.0.
\begin{table*}[h!]
\centering
\casesize
\begin{tabularx}{\textwidth}{>{\raggedright\arraybackslash}p{2.1cm} >{\raggedright\arraybackslash}X >{\raggedright\arraybackslash}X >{\raggedright\arraybackslash}X}
\toprule
 & \textbf{Iter-1} & \textbf{Iter-2} & \textbf{Iter-3} \\ 
\midrule
\textbf{Prefix} & <think>\verb|\n|\textcolor{myorange}{In summary, having completed all steps, here’s my concluding result.}  & <think>\verb|\n|\textcolor{mygreen}{Before finalizing my response, it's crucial to check the whole reasoning process for any slip-ups or oversights.} I want to make sure everything's accurate. & <think>\verb|\n|\textcolor{mygreen}{I need to avoid any potential uncertainties or hesitations,} \textcolor{myblue}{just focus on the task and execute confidently.} \\ 

\textbf{Controlled RBs} & \textcolor{myorange}{Final Conclusion}  & \textcolor{mygreen}{Uncertainty Management} & \textcolor{myblue}{Stepwise Reasoning}; \textcolor{mygreen}{Uncertainty Management} \\ 

\textbf{Score} & 59.3  & 60.2 & 63.0 \\ 
\bottomrule
\end{tabularx}
\caption{A case study on ThinkPilot's iterative optimization, detailing the prefixes, guided reasoning behaviors (Controlled RBs), and scores for three iterations in R1-Qwen-7B on IFEval benchmark. Another case study for reasoning benchmark see Appendix~\ref{app:more case}.}
\label{The case study}
\end{table*}

\begin{table}[H]
\centering
\small
\begin{tabularx}{\textwidth}{>{\raggedright\arraybackslash}p{1.5cm} >{\raggedright\arraybackslash}X >{\raggedright\arraybackslash}X >{\raggedright\arraybackslash}X}
\toprule
\textbf{Iteration} & \textbf{Iter-1} & \textbf{Iter-2} & \textbf{Iter-3} \\ 
\midrule
\textbf{Prefix} & <think>\textbackslash{}n\textcolor{violet}{I need to keep my knowledge base active. During the problem-solving process, I will actively retrieve and list all potentially useful formulas and concepts to have them ready for use.} & <think>\textbackslash{}n\textcolor{myblue}{Okay, I’m going to dive right into solving this without spending too much time identifying the detailed objectives or constraints.} \textcolor{pink}{I don’t want to overthink what exactly the task involves—better to just get moving.} & <think>\textbackslash{}n\textcolor{cyan}{Alright, I need to start solving this problem. Let me just jump right into executing the steps without spending too much time planning it all out.} \\ 
\addlinespace

\textbf{Controlled RBs} & \textcolor{violet}{Knowledge Retrieval } & \textcolor{pink}{Stepwise Reasoning;  } 

\textcolor{myblue}{Task Initialization} & \textcolor{cyan}{Strategic Planning} \\ 
\addlinespace

\textbf{Score} & 50.0  & 56.7 & 60.0 \\ 
\addlinespace
\bottomrule
\end{tabularx}
\caption{A case study on ThinkPilot's iterative optimization, detailing the prefixes, guided reasoning behaviors (Controlled RBs), and scores for three iterations.}
\label{The case study of appendix}
\end{table}


\section{More Related Work}
\paragraph{Large Reasoning Models}

Recent large reasoning models~\citep{jaech2024openai, guo2025deepseek,yang2025qwen3} use intermediate steps, known as Chain-of-Thought (CoT) \citep{wei2022cot}, to tackle complex problems more effectively. Extensions like multi-path sampling \citep{wang2022self}, trees \citep{yao2023tree}, and graphs \citep{besta2024graph} enhance this further. However, these self-generated processes often lack control, leading to verbosity and poor instruction adherence—highlighting the need for methods that can effectively guide the model's reasoning.

\section{Ethics, Broader Impact, and Licenses}

\noindent
\textbf{Ethics.} A portion of our research is dedicated to the responsible and ethical development of AI systems, including but not limited to improving the safety and alignment of reasoning models. We emphasize instruction following across a variety of real-world scenarios, particularly those with heightened requirements for reliability and ethical standards. Our methodology adheres to widely accepted ethical standards in AI research, prioritizing transparency and minimizing potential societal harms. In evaluating model performance on safety-related benchmarks, we have used datasets that may contain sensitive content. All such datasets are sourced from reputable and reliable providers, ensuring research integrity and ethical compliance. Although some datasets include sensitive material, their use is strictly limited to academic research purposes and is carefully managed under controlled and ethical conditions.

\vspace{6pt}
\noindent
\textbf{Broader Impact.} Improving the safety alignment of language models in reasoning tasks has significant potential for positive outcomes in high-impact domains such as healthcare, finance, and education. At the same time, we recognize and take seriously the risks associated with misuse or unintended consequences when deploying these models. We encourage proactive research and regulatory measures to identify, monitor, and mitigate such risks.

\vspace{6pt}
\noindent
\textbf{Licenses.} In this paper, we utilize the following models and datasets: (1) \textbf{Models:} DeepSeek-R1-Distill-Qwen-1.5B (Apache 2.0 License), Qwen3-8B (Apache 2.0 License), QwQ-32B (Apache 2.0 License), THINKPRUNE (Apache 2.0 License), \citet{arora2025training} (Apache 2.0 License), DeepSeek-R1-Distill-7B (Apache 2.0 License), DeepSeek-R1-Distill-32B (Apache 2.0 License), SAFECHAIN (Apache 2.0 License), and RealSafe-R1 (Apache 2.0 License). (2) \textbf{Datasets:} XSTest (Attribution 4.0 International), StrongREJECT (MIT License), IFEval (Apache 2.0 License), MultiChallenge (Not specified), MATH 500 (MIT License), AIME 2024 (Not specified), GPQA-D (MIT License), AMC 2023 (Not specified), GPQA\_main (MIT License), AMC 2022 (Not specified), and AIME 2023 (Not specified). For more details about the licenses and usage permissions, please refer to the official documentation of each model and dataset.

\end{document}